\documentclass[letterpaper, 10 pt, journal, twoside]{IEEEtran}

\IEEEoverridecommandlockouts                              % This command is only needed if 
\usepackage{cite}
\usepackage{balance}
\usepackage{graphicx}
\usepackage{amsmath} 		% assumes amsmath package installed
\usepackage{amssymb}  	% assumes amsmath package installed
\usepackage{dsfont}
\usepackage{mathrsfs}
\usepackage[usenames,dvipsnames]{color}
\usepackage{subfig}
\usepackage{forloop}
\usepackage{ifthen}
\usepackage{intcalc}
\usepackage{tikz}
\usepackage{textcomp}
\usepackage{float}
\usepackage{amsmath,scalerel}
\usepackage{bm}
\usepackage[ruled,vlined,linesnumbered]{algorithm2e}
\usepackage{multirow}
\newcommand {\mibf}[1] {\boldsymbol{#1}}

\newcommand {\emilista} {\end{list}}
\newsavebox{\ieeealgbox}

\begin{document}

\title{\LARGE \bf
%Knowledge-oriented Physics-based Motion Planning for Grasping in Uncertain Environment
Randomized Physics-based Motion Planning \\for Grasping in Cluttered and Uncertain Environments}

\author{Muhayyuddin$^1$,~%~\IEEEmembership{Member,~IEEE,}
        Mark Moll$^2$,~%~\IEEEmembership{Fellow,~OSA,}
        Lydia Kavraki$^2$,~%~\IEEEmembership{Fellow,~OSA,}
        Jan~Rosell$^1$%,~\IEEEmembership{Life~Fellow,~IEEE}% <-this % stops a space
\thanks{$^1$Universitat Polit\`ecnica de Catalunya, Institute of Industrial and Control Engineering, {\ttfamily \{muhayyuddin.gillani, jan.rosell\} at upc.edu}. Work on this paper was sponsored in 
part by Spanish Government project \mbox{DPI2016-80077-R}.}% <-this % stops a space
\thanks{$^2$Rice University, Department of Computer Science, Houston, TX 77005, {\ttfamily \{mmoll, kavraki\} at rice.edu}. Work on this paper was sponsored in 
part by  NSF IIS1317849 and IIS1718478}
}

\markboth{IEEE ROBOTICS AND AUTOMATION LETTERS. PREPRINT VERSION. ACCEPTED NOVEMBER, 2017}%
{Muhayyuddin \MakeLowercase{\textit{et al.}}: Randomized Physics-based Motion Planning for Grasping in Cluttered and Uncertain Environments}

\maketitle
%\thispagestyle{empty}
%\pagestyle{empty}

%%%%%%%%%%%%%%%%%%%%%%%%%%%%%%%%%%%%%%%%%%%%%%%%%%%%%%%%%%%%%%%%%%%%%%%%%%%%%%%%
\begin{abstract}
Planning motions to grasp an object in cluttered and uncertain environments is a challenging task, particularly when a collision-free trajectory does not exist and objects obstructing the way are 
required to be carefully grasped and moved out. This paper takes a different approach and proposes to address this problem by using a randomized physics-based motion planner that permits robot-object 
and object-object interactions. The main idea is to avoid an explicit high-level reasoning of the task by providing the motion planner with a physics engine to evaluate possible complex multi-body 
dynamical interactions. The approach is able to solve the problem in complex scenarios, also considering uncertainty in the objects' pose and in the contact dynamics. The work enhances the state 
validity checker, the control sampler and the tree exploration strategy of a kinodynamic motion planner called KPIECE. The enhanced algorithm, called p-KPIECE,  has been validated in simulation and 
with real experiments. The results have been compared with an ontological physics-based motion planner and with task and motion planning approaches, resulting in a significant improvement in terms of 
planning time, success rate and quality of the solution path.
\end{abstract}

\section{Introduction} 

\IEEEPARstart{I}{n} the past decades, motion planning has been considered under deterministic conditions, such as the positions of the objects and the motion of the robots being precisely 
defined. Recently, the incorporation of uncertainties and probabilistic treatment of the planning problem has gained a lot of interest, with the objective to compute robust motion plans for real 
environments. In this line, planning the grasping motions in unstructured and uncertain environments is a challenging task, particularly when a collision-free trajectory from start to goal does not 
exist. To tackle this issue, two approaches are mainly followed: \textit{task and motion planning}~\cite{dantam2016,srivastava2014,hadfield2015,stilman2005,stilman2007} and \textit{clutter 
grasping}~\cite{dogar2010,dogar2011,dogar2012,LaskeyEtAl2016b}. The former breaks down the problem into a repeated sequence of actions to remove the objects obstructing the path towards the target: 
select an object to move, select the grasping points (or contact points to push) on the object, compute a collision-free trajectory for grasping (pushing), and find an appropriate 
position to place the grasped (pushed) object. Finally, a collision-free trajectory is computed to grasp the target object. On the other hand, clutter grasping, the topic of this paper, 
is based on planning motions such that interactions with the objects are allowed, in order to clear the path to grasp the target object.
 
The \textit{task and motion planning} approaches work well for structured or semi-structured environments. However, cluttered and uncertain environments raise challenging issues. On the one hand, these approaches require the detailed semantic description of the scene and the explicit reasoning about each robot-object interaction (to carefully move objects in clutter). On the other hand, uncertainty has to be also considered at task level regarding the action effects, which leads to computationally intensive methods that may fail in highly cluttered and uncertain environments, i.e., it may not be possible to find a robust sequence of actions to free the path towards the target object.
These challenging issues could be tackled more easily using \textit{clutter grasping} approaches. These strategies free the path towards the target object by pushing the objects away without explicitly reasoning about each interaction. 
%A few approaches, such as \cite{dogar2010}, have worked in this direction by considering several straight-line motions and selecting the one that ended up at a pre-grasping position (modelling interactions with a quasi-static assumption). These strategies may fail if no solution exists with straight-line motions. Moreover, the complex dynamical robot-object and object-object interactions are difficult to model accurately, and therefore, if not correctly done, can give rise to non-realistic results. To handle uncertainty, \textit{clutter grasping} approaches usually compute the robot-object interactions at a preprocessing step, where any potential deviation in the objects motions in the result of interactions is determined. Afterwards, the potential deviation is considered in the planning process. 
This paper proposes to address the challenging issues of clutter grasping by using a randomized physics-based motion planner for grasping in the presence of uncertainty. The planner is not constrained to straight-line motions and  does not require any preprocessing step.  

\textit{Contributions:} The main contribution of this paper is the proposal of a physics-based motion planning strategy (framed within sampling-based motion planning strategies) for grasping in cluttered and uncertain environments. The main components of the paper are \textit{(1)} \textit{A probabilistic control sampler} that samples controls and computes the belief about the validity and the robustness of the sampled controls in the presence of object pose uncertainty; (2) \textit{A tree exploration strategy} which integrates the computed belief with the planning data structures, biasing the exploration towards the states that have high belief; \hbox{(3) \textit{An uncertainty handling strategy}} to cope with the  propagation to future states of the objects'  pose uncertainty which arises from robot-object or object-object interactions. To handle the environment dynamics, Open Dynamic Engine (ODE, {\sf \small www.ode.org}) is used as state propagator. The proposal has been implemented as a variant of the KPIECE planner~\cite{sucan2012}. 

The rest of the paper is structured as follows. First, Sec.~\ref{sec:Rwork} provides an overview of the relevant literature, and Sec.~\ref{sec:pformulation} formulates the problem and provides an overview of the solution. Then, Sec.~\ref{sec:propoal} explains the proposed approach along with the algorithms, and Sec.~\ref{sec:Evaluation} describes the simulation set-up and compares the proposed approach with others. Finally, Sec.~\ref{sec:conclusion} discusses the results of the work.

\section{Related Work}\label{sec:Rwork}

In a manipulation task, the main sources of uncertainty are the imprecise knowledge of the initial state of the system (sensing uncertainty), of the robot dynamics (motion uncertainty) and of the future state of the environment when interactions exist (environment uncertainty). 
Sensing uncertainty is tackled, for instance, in~\cite{miralles2004} that describes the obstacle map using Gaussian distributions, and constructs a road-map with minimum probability of collision. Motion uncertainty is considered in different ways.  Some approaches~\cite{Bry2011} consider the uncertainty in the robot dynamics and in the initial state of the system as a zero mean Gaussian noise, and use a variant of RRT~\cite{lavalle2001} that, at each step, computes the distribution of the state  in a way such that the robot keeps a safe distance from the obstacles. Others use a linear quadratic regulator to maximize the probability to reach the goal or to minimize an expected cost function~\cite{van2011}, or rely on 
\textit{Markov Decision Processes} (MDPs) to compute the global control policy over the environment to maximize the probability of success~\cite{alterovitz2007}. Uncertainty in the environment is 
considered in~\cite{Melchior2007}, which treats the extension step of an RRT as a stochastic process, simulating multiple times and using clustering techniques to create nodes, thus finding  
inherently safer paths. Within the scope of RRT planners, a metric is introduced in~\cite{JohnsonKS2016} to guide the search towards more convergent trajectories, which increases robustness, as 
demonstrated with a manipulator rearranging objects. Uncertainty in rearrangement problems is also tackled in~\cite{KingRS2017} using a learned policy from user demonstrations, and with a similar idea 
in~\cite{LaskeyEtAl2016b} for grasping in clutter. Learning from demonstrations avoid assuming explicit knowledge of object or dynamics models, although this has been done to include uncertainty 
while planning for clutter grasping, assuming straight-line motions and quasi-static push mechanics~\cite{dogar2010}\cite{dogar2012}.

Physics-based motion planning is a step further towards physical realism in the planning process, and therefore useful in manipulation tasks. It extends kinodynamic motion planners (i.e., those 
planners that consider kinematic and dynamics constraints) by incorporating the physics-based constraints, and by allowing the robot-object and object-object interactions during planning. These 
interactions are modeled based on rigid body dynamics. Typically, sampling-based kindoynamic motion planners (particularly tree-based planners) are used to sample the states and to construct the 
solution path. The propagation step is performed using a physics engine such as ODE. Within this framework, approaches are proposed for problems like motion planning among collisionable 
obstacles~\cite{Muhayyuddin2017}, rearrangement planning~\cite{Haustein20015}~\cite{KingCS2016}, object placement~\cite{CosgunHES2011}, object sorting~\cite{GuptaMS2015} or bin 
picking~\cite{MahlerG2017}. This paper will tackle the clutter grasping problem using a physics-based motion planner able to cope with uncertainty in the initial state of the system and in the poses 
of the objects as a result of robot-object or object-object interactions. 
 
\section{Problem Formulation}\label{sec:pformulation}

\subsection{Problem Statement}\label{sec:pstatement}
Consider a motion planning problem that asks a robot to grasp an object in a cluttered and uncertain environment, where possibly no collision-free trajectory exists that moves the robot from 
an initial configuration to a pre-grasping pose. All the objects are assumed to have stable support surfaces and be laying on a flat working region (e.g., a table). The objective is to compute the 
sequence of robust controls and their durations in such a way that, if applied to the system in the presence of object pose uncertainty and imprecise knowledge of contact dynamics, it moves the robot 
from the start to the goal state (a pre-grasping pose), by pushing away those moveable objects obstructing the path, while avoiding collisions with fixed obstacles. Object-object interactions 
between 
moveable objects are 
allowed. Note that we are not dealing with the rearrangement problem, i.e. the final pose of the objects is not relevant.

Moreover, in order to ease the finding of a solution, the following constraints are set: 1) no interaction is allowed with the target object; 2)  as a result of interactions the object's 
velocity 
must be less than a given threshold, and no object should change its support surface or fall from the working region.

\subsection{World Modeling and State Validation}\label{sec:wmodel}
Let $\mathcal{X}$ be a differential manifold representing the state space of the robot. At any time $t$ the state of the robot $x_t\in \mathcal{X}$
is specified as $x_t=(\mibf{q}_t,\mibf{\dot{q}}_t)$, where $\mibf{q}_t$ and $\mibf{\dot{q}}_t$ describes the configuration of the robot and its time derivative, respectively. The objects in the 
workspace are classified as fixed, movable and target object, and are represented as $\mathcal{O} = 
\{\mathcal{O}^\textrm{target},\mathcal{O}^\textrm{movable}_1,\dots,\mathcal{O}^\textrm{movable}_\textrm{M},\mathcal{O}^\textrm{fixed}_1,\dots,\mathcal{O}^\textrm{fixed}_\textrm{F}\}$ with M and F 
being the number of movable and fixed objects, respectively.   Their state space $\mathcal{S}$ consists of two components, \hbox{$\mathcal{S}=\{\mibf{\rho},\mibf{v}\}$} with $\mibf{\rho}$ representing 
the pose, and $\mibf{v}$ the linear and angular velocities. At any time $t$ the state space $s_t \in \mathcal{S}$ of the target and movable objects is represented as 
\hbox{$s_t=\{s_\textrm{target},s_1,\dots,s_\textrm{M}\}$}. The state of the world is modelled as $\mathcal{W}=\mathcal{X}\times\mathcal{S}$.

The discrete-time dynamic model of the robot is:
\begin{equation}\label{eq1}
 \mathcal{W}_{t+1}= \mibf{f}(\mathcal{W}_t,\mibf{u}_t),
\end{equation}
where \hbox{$\mibf{f}:\mathcal{W}\times\mathcal{U}\rightarrow\mathcal{W}$} is the state transition function, with control space $\mathcal{U}$ representing all possible
controls that can be applied to the system, and $\mibf{u}_t\in\mathcal{U}$ a control input that is applied at time $t$.

Let $\mathcal{F}$ be a state validity checker defined as \hbox{$\mathcal{F}:\mathcal{W}\times\mibf{C} \rightarrow \{ {\sf\small true}, {\sf\small false}\}$} where $\mibf{C}$ represents the set of 
validity constraints $\mibf{C}=\{kinodynamic, interaction\}$. The $kinodynamic$ constraints refer to the workspace bounds, the joints limits, bounds over the velocities, forces, and torques. The 
$interaction$ constraints refer to the set of constraints that describe how the robot can interact with the environment, as described in the previous subsection.
A state is said to be valid if it satisfies all these validity constraints.

\subsection{Modeling Uncertainty}\label{sec:muncertainty}
The current work handles: 1) uncertainty in the initial state of the environment, 2) uncertainty in the robot motions and 3) uncertainty due to dynamic interactions. 
%%%\notas{The planner in this work is general, It can work with any arbitrary definition of uncertainty. For simplicity we are modeling uncertainty using Gaussian.}

%\begin{enumerate}
 % \item 
 1) The uncertainty in the initial pose of the objects can be attributed  to the process noise in the sensing of the environment as measured, for instance, with a RGBD camera. In this work 
object pose uncertainty is modeled as:
\begin{equation}\label{eq:urgn}
\mibf{U}_{\textrm{init}} 
\sim \mathcal{N}(\mibf{\rho}^\textrm{init},\mibf{m}^\textrm{init}),
\end{equation}
where $\mathcal{N}$ represents the multivariate Gaussian distribution with $\mibf{\rho}^\textrm{init}$ being the mean representing the set of 
measured initial poses of the objects, \hbox{$\mibf{\rho}^\textrm{init}=\{\rho^\textrm{init}_1 \dots \rho^\textrm{init}_\textrm{M}\}$}, and $\mibf{m}^\textrm{init}$ the variance,  
$\mibf{m}^\textrm{init}=\{m^\textrm{init}_1\dots m^\textrm{init}_\textrm{M}\}$. 
Other uncertainty models could be alternatively used as needed.

%\item
2) The stochastic discrete-time dynamic model of the robot is described as:
\begin{equation}\label{eq4}
 \mathcal{W}_{t+1}= \mibf{g}(\mathcal{W}_t,\mibf{u}_t,\mibf{\varepsilon}_t),
\end{equation}
where \hbox{$\mibf{g}:\mathcal{W}\times\mathcal{U}\times\mibf{\xi}\rightarrow\mathcal{W}$} is the state transition function, with the disturbance vector space $\mibf{\xi}$ containing 
all possible disturbances that can be applied to the system. Disturbance in the robot controls is modelled as:
\begin{equation}\label{eq5}
\mibf{U}_\varepsilon\sim \mathcal{N}(\mibf{0},\mibf{m}_\varepsilon). 
\end{equation}
The disturbance vector $\mibf{\varepsilon}_t \in \mibf{\xi}$ represents the control uncertainty at time $t$, and is used to introduce uncertainty in the control input at this instant of 
time.

%\item
3) The uncertainty in the future states is propagated when robot-object and object-object dynamic interactions occur. These dynamic interactions involve various dynamic parameters (such as 
friction, the pressure distribution under the object surface, the contact forces and the inertial effects), whose values greatly influence the behavior of the system.
 Physics engines provide a good approximation of the actual contact dynamics. ODE, in particular, does it with three contact parameters $\mathcal{D}=\{\mu,c,e\}$, with  $\mu$ being the vector of 
friction coefficients between pairs of objects, $c$ a constraint force mixing parameter to soften constraints and $e$ an error reduction parameter to relax the constraints satisfaction.
The uncertainty in 
the interactions will be modelled by setting a multivariate Gaussian distribution around the predefined approximated values:
\begin{equation}\label{eq3}
 \mibf{U}_{\mathcal{D} } \sim \mathcal{N}(\mathcal{D}^\textrm{aprox},\mibf{m}_{\mathcal{D}}),
\end{equation}
with mean values  $\mathcal{D}^\textrm{aprox}=\{\mibf{\mu}^\textrm{aprox},c^\textrm{aprox},e^\textrm{aprox}\}$ and variance  $\mibf{m}_\mathcal{D}=\{m_{\mu},m_{c},m_{e}\}$. 
The values of these parameters  are tuned in simulation and are qualitatively evaluated; the associated variances are small. 

%\end{enumerate}

 \begin{figure}
\begin{center}
   \includegraphics[width=1\linewidth]{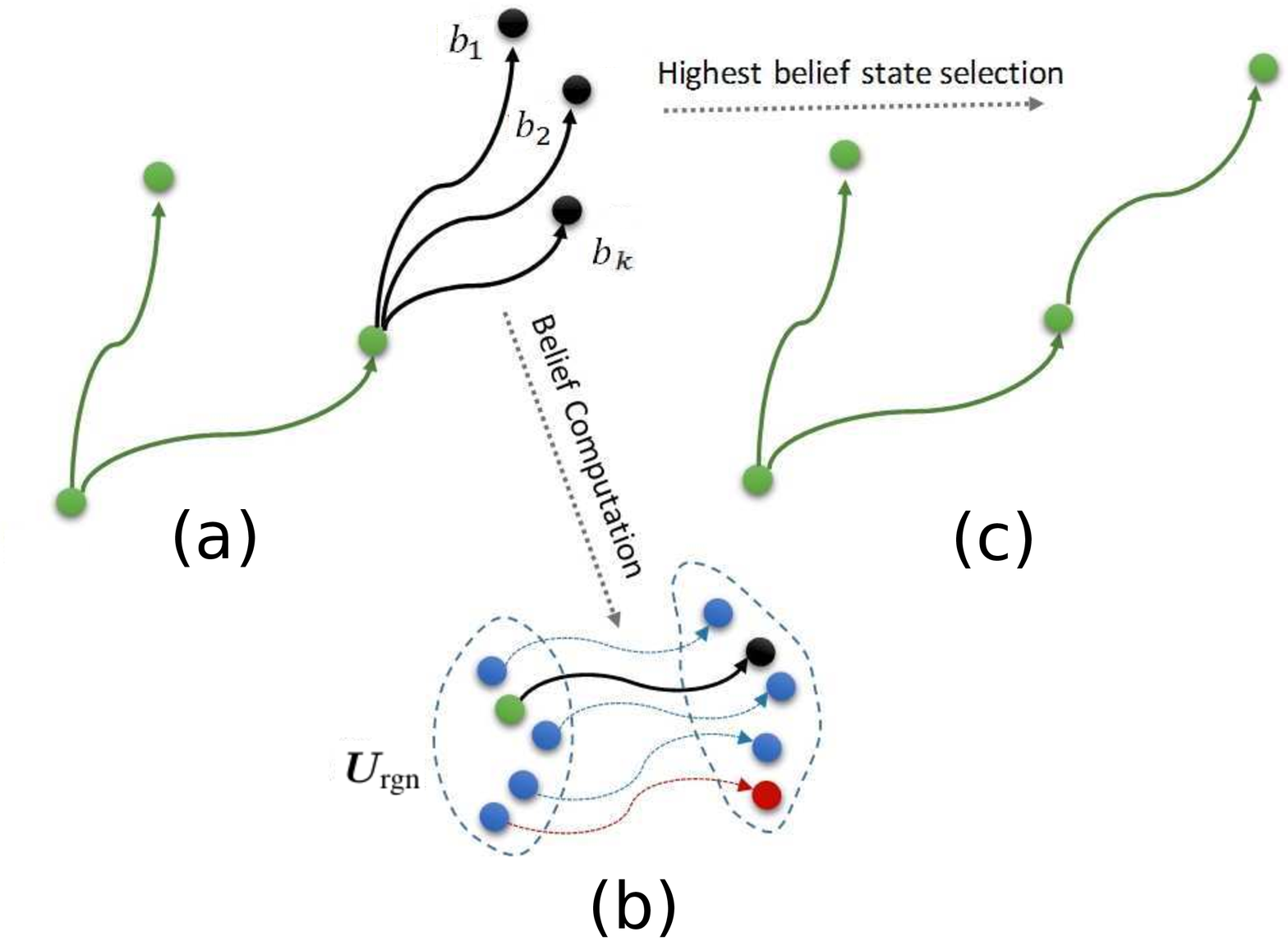}
   \caption{Control selection process. The motions in green belong to the tree, motions in black are the candidate motions and the motions in blue are the particle motions that are used to compute the 
belief of the candidate motions.}
   \label{fig:sampling}
\end{center}
\end{figure}

\section{Proposed Approach}\label{sec:propoal}
\subsection{Solution Overview}\label{sec:soverview}
The proposed approach considers the above stated motion planning problem as an open-loop problem, i.e., it tries to develop a robust strategy that absorbs the potential deviations in the objects poses and in the results of the interactions occurring during the planning process. The work presents a physics-based motion planner with an underlying sampling-based kinodynamic planner that incorporates a robust tree growing strategy. This strategy works in two phases. The first one is a control sampling phase, once the state from which the tree will grow is selected, the algorithm applies randomly sampled controls called candidate controls for some randomly sampled time durations to generate valid motions called candidate motions (as depicted in \hbox{Fig.~\ref{fig:sampling}-a)}. No uncertainty is considered in this phase. The second phase  considers the uncertainty in the system, as stated in \hbox{Sec.~\ref{sec:muncertainty}}, to compute the belief about the robustness and validity of the candidate motions. This is done for 
each candidate motion by considering the uncertainty and repeatedly applying the candidate control to obtain a set of associated motions, called particle motions (as shown in 
Fig.~\ref{fig:sampling}-b, where $\mibf{U}_{\textrm{rgn}}$ represents the uncertainty in the object's pose as introduced in Sec.~\ref{urgn}), and evaluating their validity.  This belief is used to select the best candidate motion (Fig.~\ref{fig:sampling}-c), thus enhancing the tree exploration process. Additionally, a strategy to handle the uncertainty due to dynamic interaction is included in the approach. 

%\subsection{Assumptions}
%We assume that the uncertainty due to interaction is independent from the uncertainty in the previous states.

\subsection{Kinodynamic Motion Planning}
Kinodynamic Motion Planning by Interior and Exterior Cell Exploration (KPIECE)~\cite{csucan2009, sucan2012} is used to plan the trajectory. It is a sampling-based kinodynamic motion planner, 
particularly designed 
for systems with complex dynamics. A recent benchmarking study~\cite{gillani2016} concluded that KPIECE is a very good choice (to plan efficiently) for physics-based motion planing.  
%It shows significant improvement (in terms of planning time and memory) over other sampling based kinodynamic planners such as RRT~\cite{lavalle2001}. 

KPIECE grows a tree of motions. A motion is defined as $\nu=(\mibf{x},\mibf{u},d)$, where $\mibf{x}$ is the start state of the motion, $\mibf{u}$ is the control vector that is applied at state $\mibf{x}$ for a time duration~$d$. A key feature of KPIECE is how it estimates coverage during exploration. In order to estimate the coverage, the state space is projected into a low-dimensional space that is partitioned into cells. As a result of the projection, motions 
are classified into cells (each cell can contain several motions). These cells are divided into interior and exterior, depending on whether neighboring cells are occupied or not. The tree growing procedure is based on first selecting a cell, then selecting a motion of that cell and a state pertaining to it, from where the tree will grow. The cell selection process for the tree growth is based on a parameter called \textit{importance} (cells with high importance are preferred for expansion), defined for a cell $z$ as: 
\begin{equation}\label{eq:imp}
 \textrm{Imp($z$)}=\frac{\log(\mathcal{I}) \cdot score}{C_{\textrm{cell}} \cdot (1+|Neigh(z)|) \cdot Cov(z)}\textrm{,}
\end{equation}
where $\mathcal{I}$ represents the planning iteration when the cell was added to the grid,  $C_\textrm{cell}$ represents the number of times it has been selected, $Cov(z)$ represents the number of 
states in the cell, $|Neigh(z)|$ is the number of instantiated neighboring cells, and  $score$ represents the exploration 
progress, a value that is computed by evaluating the increase in the total coverage and the time spent to achieve this increment in coverage (when expanding from the cell). Once the cell is chosen, a 
motion from that cell is selected using a half normal distribution (over the indices of the motions, ordered starting with the newest). Finally, a state is randomly picked from the selected motion and 
a new motion is sampled from that selected state. The process continues until the tree of motions reaches the goal state. 

The current proposal extends this algorithm with: 
\begin{enumerate}
\item Strategies to handle the propagation of objects' pose uncertainty into future states resulting from robot-object or object-object interactions
(Sec.~\ref{sec:upropagation}).
 \item A motion sampling strategy that samples motions to expand the tree and computes the belief about the robustness of the sampled motions \hbox{(Sec.~\ref{sec:Csampler})}.
\item A tree exploration strategy to modify the cell and motion selection process according to these beliefs (Sec.~\ref{sec:tree}).
\end{enumerate}

The algorithm, inspired by the original KPIECE algorithm, is presented in Algorithm~\ref{mainalgo}. As input it takes the initial  state of the world $\mathcal{W}_{init}$, the uncertainty in the initial pose of the objects $\mibf{U}_{\textrm{init}}$, a goal region $\mathcal{Q}_{goal}$, a time threshold $T_{max}$,  the number of candidate motions $k$ to 
use, the number of particle motions $n_p$ to evaluate the robustness, the displacement threshold $d$ for evaluating interactions,   and the validity constraints $C$ that are to be used by the state 
validity checker. It starts with $\nu_0$, a motion of zero duration representing the initial state.

Lines 1 to 3, 12 and 14 are the default steps of KPIECE. The contribution of the current proposal is implemented in function {\sf 
\small UpdatePoseUncertainty} (line-13) that updates the object pose uncertainty (Sec.~\ref{urgn}), in function {\sf \small MotionSampler} (line-8) that samples $k$ motions and select the 
one that has highest belief (Sec.~\ref{sec:Csampler}), and in functions {\sf \small SelectMotion} \hbox{(line-7)} and {\sf \small UpdateCellImportance} \hbox{(line-15)} that, respectively, select a 
motion from the selected cell to grow the tree and update the cell importance (Sec.~\ref{sec:tree}).

\begin{algorithm}
{
 \SetKwInOut{Input}{inputs}{}{}
  \SetKwInOut{Output}{output}
  \SetKwProg{ControlSampler}{ControlSampler}{}{}
  \Input{Initial State $\mathcal{W}_{init}$, initial pose uncertainty $\mibf{U}_{\textrm{init}} $, goal region $\mathcal{Q}_{goal}$, maximum time $T_{max}$, 
 candidate motions $k$, particle motion $n_p$, displacement threshold $d$, validity constraints $C$}
  \Output{A sequence of motions that move the robot to the goal region} \sf \small
   $\nu_0$ = $\mathcal{W}_\textrm{init}$\\
 $G$.empty()\\
 $G$.AddMotion($\nu_0$)\\
 $\mibf{U}_\textrm{rgn} = \mibf{U}_\textrm{init}$\\
\While{$T_{max}$ }{
 cell = SelectCell($G$)\\
 $\nu$= SelectMotion(cell)\\
 $\nu_{new}$=MotionSampler($\nu,k,n_p,d,C,\mibf{U}_\textrm{rgn}$) \\
 \If {$\nu_{new} \in Q_{goal}$}{
  \KwRet{ path to $\nu_0$}\\
  }
  \If{$\nu_{new} != $NULL}{
  $G$.AddMotion($\nu_{new}$)\\
  $\mibf{U}_\textrm{rgn} \leftarrow $UpdatePoseUncertainty($\mibf{U}_\textrm{rgn}$)\\
  }
 UpdateParameters()\\
 UpdateCellImportance(cell)
}
\KwRet {\small \sf NULL}
}%close small
\caption{Probabilistic KPIECE} \label{mainalgo}
\end{algorithm}

\subsection{Handling Uncertainty Propagation}\label{sec:upropagation}
To handle uncertainty, the most robust motion should be chosen, i.e., those with higher chance to reach a valid state. For this purpose, the robustness of the candidate motions is evaluated by generating a set of particle motions. The belief of the candidate motions is computed as a function of the validity of the motion particles. This section explains how the particle motions are generated and used to update the uncertainty pose of the objects when interactions take place.  

\subsubsection{Particle Motion Generation}
Particle motions are generated to evaluate a nominal candidate motion. Each particle motion is generated by sampling the initial state, the dynamic interaction parameters and the control to be applied: a) the initial state is sampled from the PDF function associated to the initial state of the nominal candidate motion (initially represented by Eq.~(\ref{eq:urgn})); b) the dynamic parameters are sampled using the PDF function described in Eq.~(\ref{eq3}); c) the control is determined by randomly adding noise to the nominal candidate control using Eq.~(\ref{eq5}).  Then, a particle motion is obtained by applying the selected control to the chosen initial state and considering the sampled dynamic interaction parameters. Each particle motion will be evaluated using the state validity checker $\mathcal{F}$.

\subsubsection{Updating Object Pose Uncertainty}\label{urgn}
Once interactions take place, the object pose uncertainty of the displaced objects is recomputed (objects may collide several times with the robot and/or with other objects). The poses of the 
objects as a result of interaction are highly variant depending on the contact regions and interaction force directions. 
Therefore, the resultant poses of the objects after applying several particle motions, can be modelled using the Gaussian Mixture Model (GMM), which is a convenient way to model the data that comes from different sources~\cite{PfaffPB2008}:
\begin{equation}\label{eq:GMM}
\mibf{U}_\textrm{rgn}=\sum_{j=1}^{n} \alpha_j\cdot\mathcal{N}(c_j,m_j) \; \textrm{with}\; \alpha_j\geq0 \;\textrm{,}\; \sum_{j=1}^{n} \alpha_j = 1.
\end{equation}

Eq.~(\ref{eq:GMM}) is computed using Expectation-Maximization (EM) algorithm. 
%It computes the PDF function to sample the data according to $\mibf{U}_\textrm{rgn}$. 
%Armadillo library is used compute the GMM. 

\begin{algorithm}
  \caption{MotionSampler{$(\nu,k,n_p,d,C,\mibf{U}_\textrm{rgn})$}}\label{algo:ControlSampler}
  {\sf \small
  \SetKwInOut{Input}{inputs}
  \SetKwInOut{Output}{output}
  \SetKwProg{ControlSampler}{ControlSampler}{}{}
    $\mathcal{M} \gets $SampleCandidateMotions{$(\nu, k)$}{\\
    \ForEach{$\nu_i \in \mathcal{M}$}{%
      $\mathcal{W}_c \gets \textrm{{\sf \small Propagate}}^c${$(\nu_i)$}\\
           \If{StateValidityChecker{$(\mathcal{W}_c,C)$}}{
           $\mibf{v}_\textrm{state} = \mibf{v}_\textrm{int.} = 0$\\
     \For{$j=1\; to \; n_p$}{
        $\rho \gets$GetObjectPoses($\mathcal{W}_c$)\\
        SampleObjectPose($\rho,\mibf{U}_\textrm{rgn}$)\\
        SampleDynamicParameters($\mibf{U}_\mathcal{D}$)\\
        $\mathcal{W}_p \gets \textrm{{\sf \small Propagate}}^p${($\nu_i)$}\\
        $valid_S \gets$StateValidityChecker{$(\mathcal{W}_p,C)$}\\
        $valid_I\gets$InteractionEvaluator{$(\mathcal{W}_p,disp,  d)$}\\
        \If{$valid_S$}{$\mibf{v}_\textrm{state}=\mibf{v}_\textrm{state}+1$}    
        \If{$valid_I$}{$\mibf{v}_\textrm{int.}=\mibf{v}_\textrm{int.}+1$}

	}
	$\mibf{P}^i_{state} \gets \mibf{v}_\textrm{state}/n_p$\\
	$\mibf{P}^i_{int.} \gets \mibf{v}_\textrm{int.}/n_p$\\
	$\mibf{b}_{\nu_i}\gets$ComputeBelief{$(\mibf{P}^i_{state},\mibf{P}^i_{int.})$}\\
	 AddMotionBelief{$(\nu_i,\mibf{b}_{\nu_i})$}\\
	 ExistValidMotion=$true$\\
    }
    }
    \If{ExistValidMotion}{
      \If{$\textrm{GenerateRandom(0,1)}>\textrm{bias}$}{
      	\KwRet{$\mibf{\nu}_i \gets$}SelectMotion{(max($\mibf{b}_{\mibf{\nu}_i}$))}
      }
      	\KwRet{$\mibf{\nu}_i \gets$}SelectRandom($\mathcal{M}$)
  }  
  \KwRet{NULL}
  }
  }
\end{algorithm}

\subsection{Probabilistic Motion Sampler}\label{sec:Csampler}
The function {\sf \small MotionSampler} (Line-8, Algorithm~\ref{mainalgo}) is used to select a robust motion to expand the tree. It is a probabilistic motion sampler that samples $k$ candidates, 
computes the belief abut their validity, and returns the one that has the highest belief to be added to the tree data-structure. The process is summarized in Algorithm~\ref{algo:ControlSampler} and 
uses the following functions:

\begin{itemize}
 \item {\sf \small SampleCandidateMotions:} Samples a set $\mathcal{M}$ of $k$ candidate motions  by first randomly choosing a single state from the input motion, and then by randomly sampling $k$ controls and durations.
 
 \item $\textrm{{\sf \small Propagate}}^c$: Uses Eq.~(\ref{eq1}) to compute the  state resulting from applying motion~$\nu_i$ (i.e., sampled control $\mibf{u}$ at the 
selected state $\mibf{x}$ for the time duration $d$).
 
 \item $\textrm{{\sf \small Propagate}}^p$: Uses Eq.~(\ref{eq4}) to compute the  state resulting from applying motion~$\nu_i$.

 \item {\sf \small StateValidityChecker:} Evaluates the resultant state of the world, $\mathcal{W}_c$,  using the state validity checker~$\mathcal{F}$. If $\mathcal{W}_c$ satisfies all 
the validity constraints,  $\mathcal{F}$ will return true, and false  otherwise. The validity constraints are explained in Sec.~\ref{sec:wmodel}.

\item {\sf \small GetObjectPoses}: Retrieves  the object poses from $\mathcal{W}_c$.

\item {\sf \small SampleObjectPose}: Samples the object poses using Eq.(\ref{eq:urgn}) or Eq.(\ref{eq:GMM}), depending on whether the object is still at its initial pose or has been moved due to interactions.

\item {\sf \small SampleDynamicParameters:} Samples the dynamic parameters using the probability distribution function described in Eq.(\ref{eq3}).
 
\item {\sf \small InteractionEvaluator:} Returns true if $|disp|\le d$ or false otherwise, where $disp$ is the displacement vector containing the displacement(s) covered by the object(s) in the result 
of interaction and $d$ a given threshold (experiments showed an increase in success rate if object pose uncertainty was reduced by avoiding very large displacements in highly cluttered 
scenes).

%Returns true if $|disp|\le d$ or false otherwise, where $disp$ is the displacement vector containing the displacement(s) covered by the object(s) in the result of interaction and $d$ a given threshold to prune too large resulting displacements that may increase too much the size of the \notas{\textit{object pose uncertainty}} and lead to a drop in performance.

\item {\sf \small ComputeBelief:} Compute the belief  of a motion as 
%\begin{equation}\label{eq:belief}
 \hbox{$b^i=\mibf{\textrm{P}}^i_{state}\mibf{\textrm{P}}^i_{int.}$}.
%\end{equation}
\item {\sf \small AddMotionBelief:} Add the belief into the corresponding motion data structure in $\mathcal{M}$. 

\item {\sf \small SelectMotion:} Selects the motion from $\mathcal{M}$ that has the highest belief, although with a given small probability, it is replaced by {\sf \small SelectRandom} that selects a motion randomly regardless of its belief value.
\end{itemize}

%\begin{figure}[t]
%\begin{center}
%   \includegraphics[width=0.8\linewidth]{figures/scene.eps}
%   \caption{Example scenario to validate the approach: a) the target object (wine glass) is surrounded by the movable objects (coke cans), the goal is to move the griper in pre-grasp 
%configuration by pushing the objects away. b) represents the scene with arbitrary shaped objects the goal is to move 
%the gripper to the pre-grasping configuration to grasp the green box.}\label{fig:scene}
%\end{center}
%\end{figure}

\subsection{Tree Exploration Strategy}\label{sec:tree}
 
The main parameters that control the tree exploration in KPIECE are the cell selection process and the motion selection from the selected cell. In order to enhance 
the exploration strategy to make the tree to grow through robust regions, our approach modifies these processes as follows.

\begin{itemize}  

\item  The definition of a motion is modified by incorporating the motion belief, i.e., 
%\begin{equation}\label{eq:motion}
$\mibf{\nu}=(\mathcal{W},\mibf{u},d,\mibf{b}_{\nu})$,
%\end{equation}
where $\mathcal{W}$ is the state of the world and $\mibf{b}_{\nu}$ represents the belief about the robustness of $\nu$. 

\item A belief value is associated with the cells, according to the beliefs of the motions they contain. It is computed as the mean value of the beliefs of the motions  normalized for all the cells, 
i.e.:
%\begin{eqnarray}
 %\overline{\mibf{b}}_{\nu} &=& \frac{1}{n}\sum_{\forall \nu_i \in \textrm{cell}} \mibf{b}_{\nu_i}  \\
 %\mibf{b}_{cell}                 &=& \frac{ \overline{\mibf{b}}_{\nu} }{\sum_{\forall \textrm{cell}j}  \overline{\mibf{b}}_{\nu}^j }
 %\end{eqnarray}
\begin{equation}
 \mibf{b}_{cell} = \frac{ \overline{\mibf{b}}_{\nu} }{\sum_{\forall \textrm{cell}j}  \overline{\mibf{b}}_{\nu}^j } \;\;\;\textrm{with}\;\;\; \overline{\mibf{b}}_{\nu} = \frac{1}{n}\sum_{\forall \nu_i 
\in \textrm{cell}} \mibf{b}_{\nu_i}
 \end{equation}
\item  The Imp parameter is modified by integrating $\mibf{b}_\textrm{cell}$ in order to favour those cells with higher belief (the strength of this bias being controlled by a heuristic parameter $f$,  set equal to the number of cells):

\begin{equation}\label{eq:imp}
 \textrm{Imp($z$)}=\frac{(1+f\mibf{b}_{cell})\cdot \log{(\mathcal{I})} \cdot score}{ C \cdot (1+|Neigh(z)|) \cdot Cov(z)}
\end{equation}
\end{itemize}

Then, once a cell is chosen for the exploration, the motion from that cell is selected based on $\mibf{b}_{\nu}$. If several motions have the same belief, a random motion will be selected from them. 
 In order to explore the less certain regions, with a fixed small probability, the selection 
is done as in the standard KPIECE.

%%%%%%%%%%%%%%%%%%%%%%%%%%%%%%%%%%%%%%%%%%%%%%%%%%%%%%%%%%%%%%%%%%%%%%%%%%%%%%%%

\section{Evaluation} \label{sec:Evaluation}
%concluding paragraph
The proposed approach described in the previous section results in a planner that enhances the power of KPIECE by considering uncertainty, object interactions and by favoring the tree exploration 
towards safer areas, allowing to plan motions in cluttered and uncertain environments. The main parameters of the algorithm are evaluated in this section, where comparisons with other approached are
also done.

\subsection{Simulation Setup}
The simulation setup is implemented in \textit{The Kautham Project}~\cite{Rosell2014}, a C++ based open-source tool for motion planning. It provides the flexibility to plan under geometric, kinodynamic and physics-based constraints.  It uses Open Motion Planning Library (OMPL)~\cite{sucanMK2012} as a core set of sampling-based planning algorithms. OMPL  provides the integration with the Open Dynamic Engine that can be used as state propagator to handle the physics-based constraints. A variant of KPIECE has been implemented that 
includes all the extensions described above. All experiments were run on an Intel Core i7-4500U 1.80GHz CPU with 16 GB memory.
\begin{figure}
\begin{center}
   \includegraphics[width=1\linewidth]{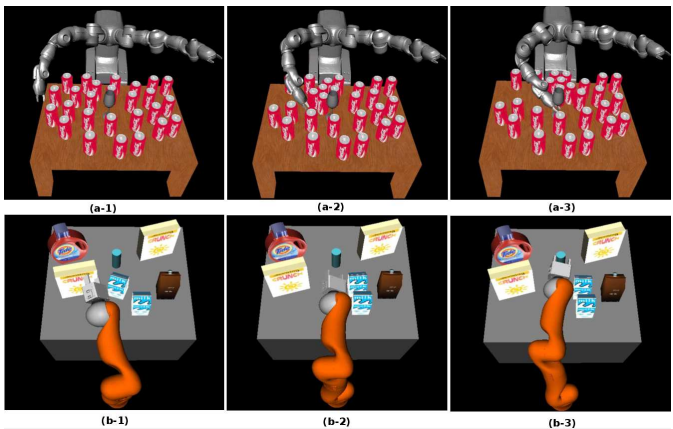}
   \caption{Sequence of the snapshots of the execution with: a) YuMi; b) Kuka-LWR. 
   Videos: https://goo.gl/RzLVi1 
   and https://goo.gl/socxhb}\label{fig:yumi-kuka}
\end{center}
\end{figure}

\begin{figure}
\begin{center}
   \includegraphics[width=1\linewidth]{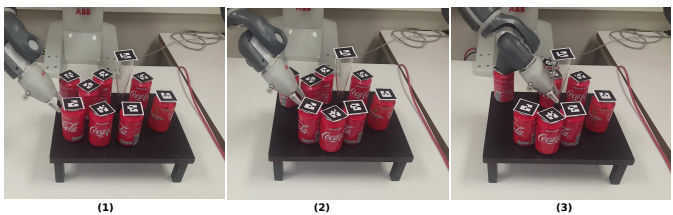}
   \caption{Sequence of snapshots of the execution on the real YuMi robot. Video: https://goo.gl/bHspjZ}\label{fig:real}
\end{center}
\end{figure}

%\begin{figure}
%\begin{center}
%   \includegraphics[width=0.9\linewidth]{figures/histogramstates.eps}
%   \caption{Histogram of average number of states and cells of 60 runs generated by o-KPIECE and p-KPIECE, respectively. }\label{fig:histstates}
%\end{center}
%\end{figure}

%\begin{figure}
%\begin{center}
%   \includegraphics[width=0.9\linewidth]{figures/planningtime.eps}
%   \caption{Graph of average planning time of 30 runs by varying the clutterness, in the environment using o-KPIECE and p-KPIECE.}\label{fig:planningtime}
%\end{center}
%\end{figure}

\begin{figure*}
\begin{center}
\begin{minipage}{0.32\linewidth}
    \centering
    \includegraphics[width=1\textwidth]{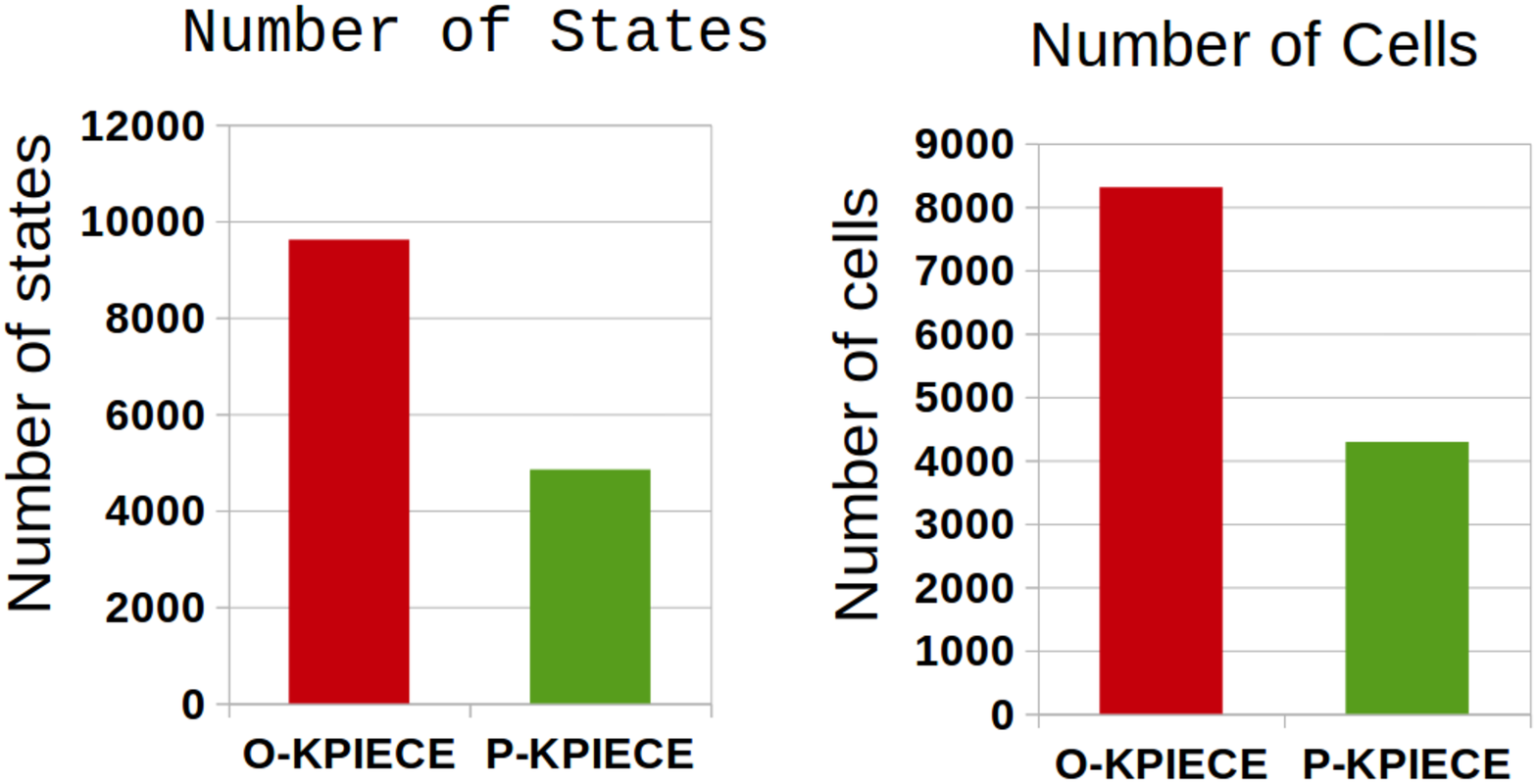}
    \caption{Histogram of average number of states and cells of 60 runs.}
    \label{fig:histstates}
\end{minipage}
\hspace{0.1cm}
\begin{minipage}{0.32\linewidth} 
    \centering
    \includegraphics[width=1\textwidth]{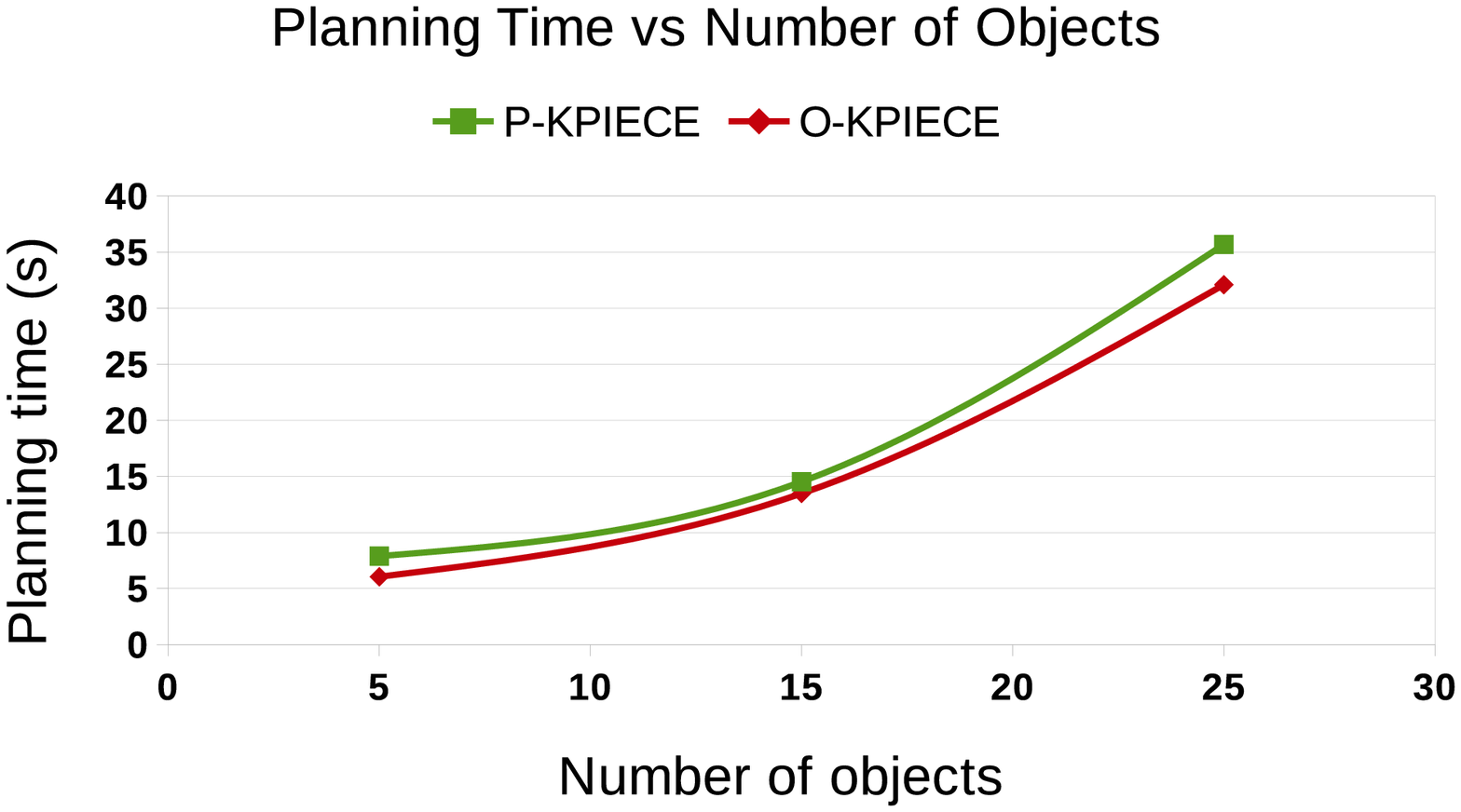}
    \caption{Graph of average planning time of 30 runs by varying the clutterness.}
    \label{fig:planningtime}
\end{minipage} 
\hspace{0.1cm}
\begin{minipage}{0.32\linewidth} 
    \centering
    \includegraphics[width=1\textwidth]{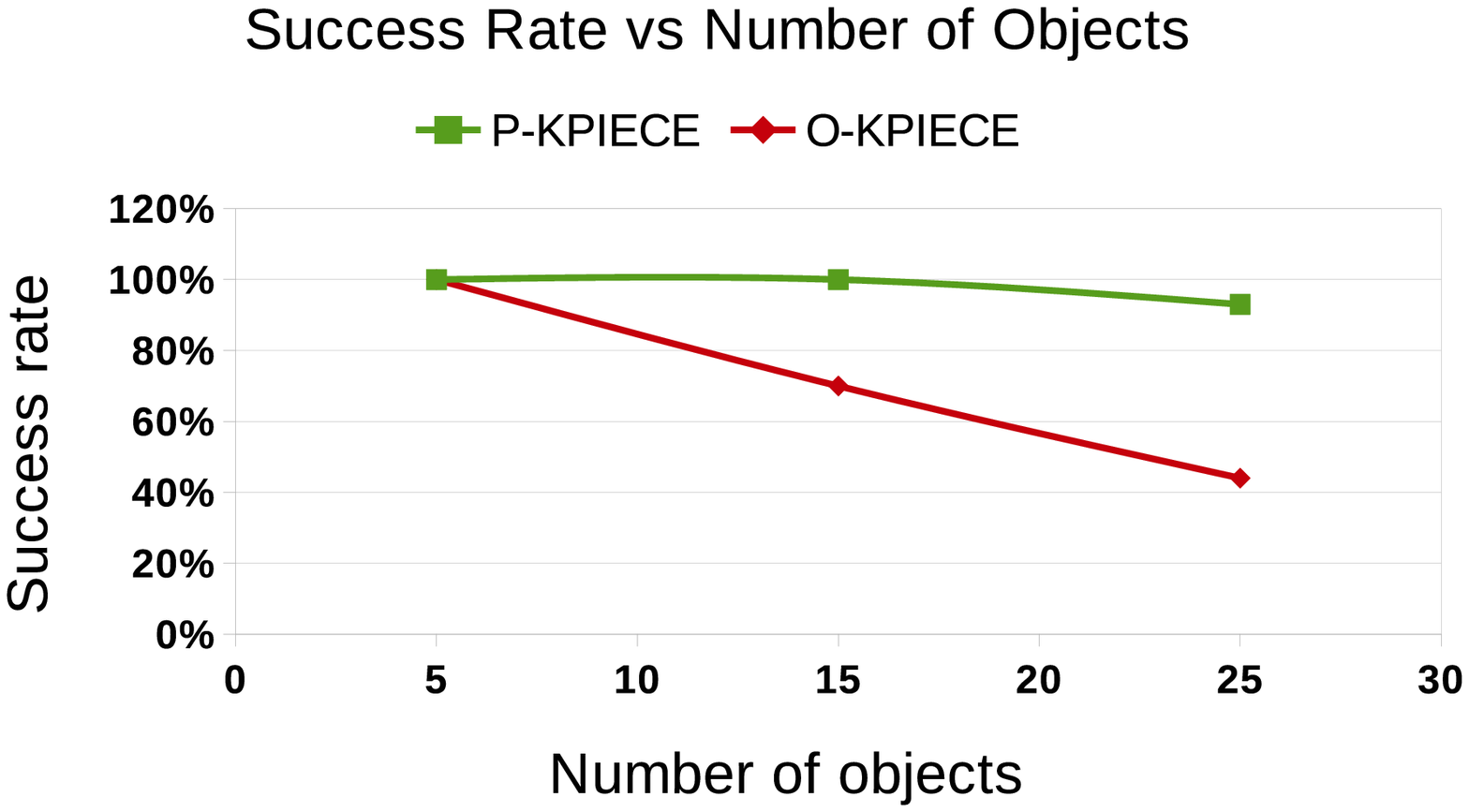}
    \caption{Graph of average planning success rate of 30 runs by varying the clutterness.}
    \label{fig:successrate}
\end{minipage}
\hspace{0.1cm}
\begin{minipage}{0.32\linewidth}
    \centering
    \includegraphics[width=1\textwidth]{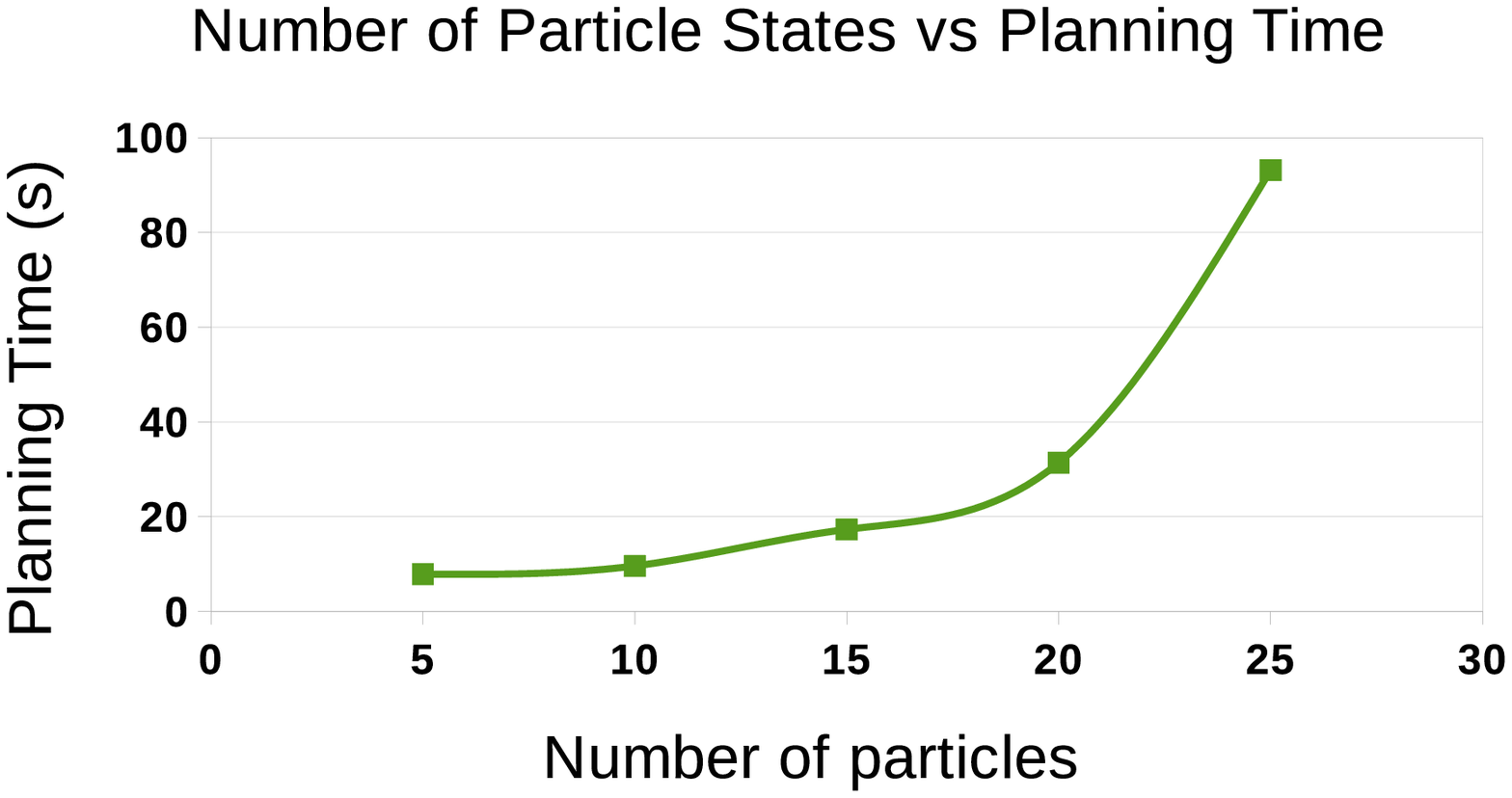}
    \caption{Graph of average planning time of 60 runs by varying the number of particle motions.}
    \label{fig:particlesplanning}
\end{minipage}
\hspace{0.1cm}
\begin{minipage}{0.32\linewidth} 
    \centering
    \includegraphics[width=1\textwidth]{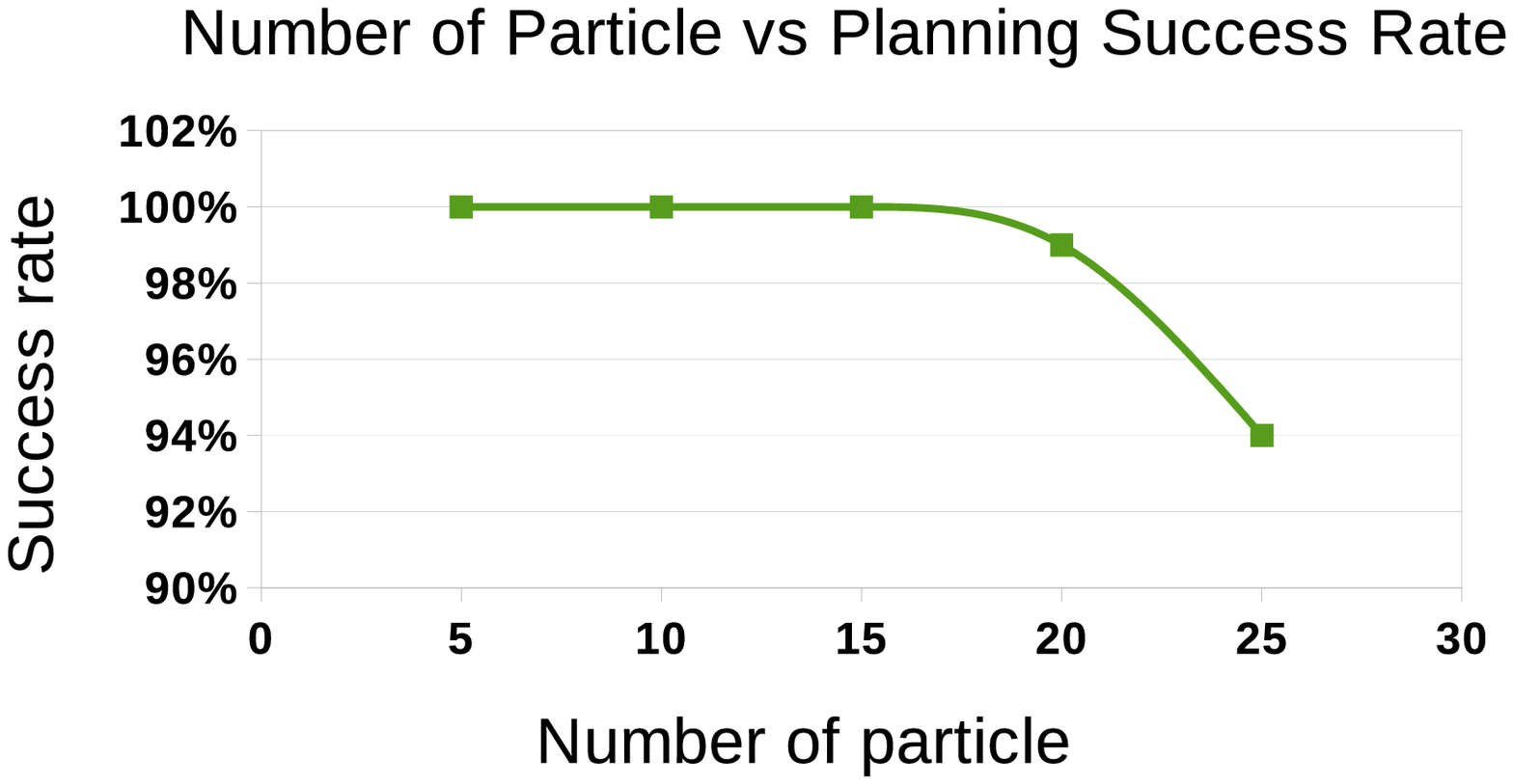}
    \caption{Graph of average planning success rate of 60 runs by varying the number of particle motions.}
    \label{fig:planningsuccess}
\end{minipage} 
\hspace{0.1cm}
\begin{minipage}{0.32\linewidth} 
    \centering
    \includegraphics[width=1\textwidth]{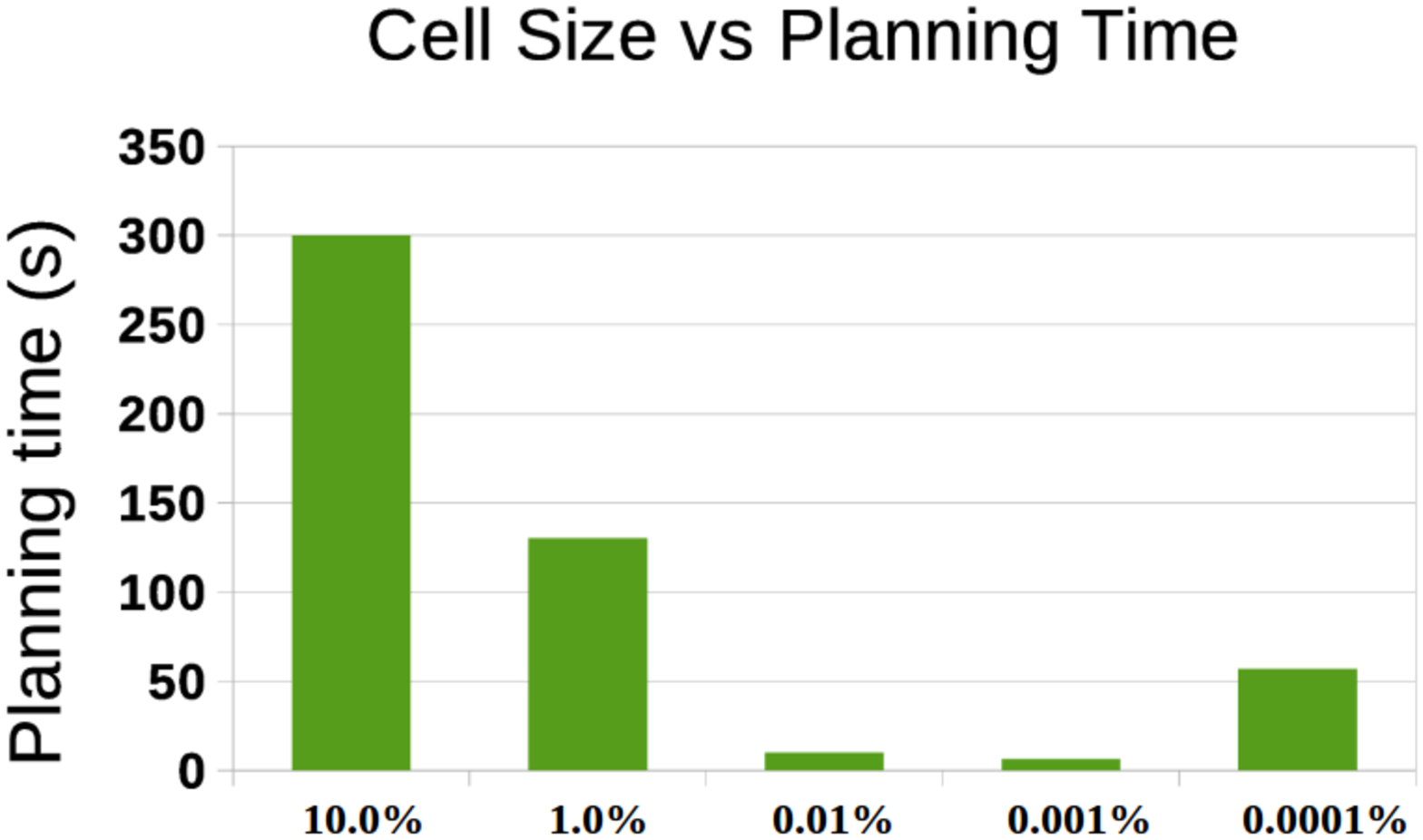}
    \caption{Histogram of average planning time of 60 runs by varying the cell size.}
    \label{fig:cellsize}
\end{minipage}
\end{center}
\end{figure*}  

The proposed approach has been tested with two different robots: the 14 degree-of-freedoms (DOF) ABB YuMi and the 7 DOF KUKA-LWR, both of them with 2-finger grippers. The example scenarios are 
presented in Fig.~\ref{fig:yumi-kuka}. The scene shown in Fig~\ref{fig:yumi-kuka}-a consists of movable objects (red cans) and the target object (wine glass). The goal is to move the robot to a 
pre-grasp 
configuration to grasp the wine glass, by pushing away those cans obstructing the path. The scene depicted in Fig~\ref{fig:yumi-kuka}-b consists of  movable objects of arbitrary shapes (boxes and 
bottle) 
and the target object (cyan cylinder). 
%The snapshot of the sequences of executions with both robots (and 
%randomly setting the initial poses within their uncertainty ranges) are depicted in Fig.~\ref{fig:yumi} and Fig.~\ref{fig:kuka}, respectively. 
Moreover, the approach is also validated with the real 
YuMi robot, and a sequence of snapshots of the execution is  depicted in Fig.~\ref{fig:real}.

%\begin{figure}
%\begin{center}
%   \includegraphics[width=0.9\linewidth]{figures/successrate.eps}
%   \caption{Graph of average planning success rate of 30 runs by varying the clutterness, using o-KPIECE and p-KPIECE.}\label{fig:successrate}
%\end{center}
%\end{figure}
\subsection{Benchmarking}
 The current approach is benchmarked against two approaches: 1) an ontological physics-based motion planner, introduced in~\cite{Muhayyuddin2015}, that enhances  KPIECE for physics-based motion 
planning 
(it will be referred as o-KPIECE); 2) a task and motion planning approach for grasping in clutter proposed in~\cite{srivastava2014,hadfield2015}. We run 30 simulation for each 
scenario.
 
\subsubsection{Algorithmic Parameter Evaluation}
 The current approach always generates more efficient solutions in terms of memory. Extensive tests showed that the number of generated states in p-KPIECE is always less than o-KPIECE. This implies that the number of generated cells is also less. Since the cell parameters are updated after each propagation step, \hbox{p-KPIECE} will update them faster, as compared to \hbox{o-KPIECE}.  Fig.~\ref{fig:histstates} shows the histograms of the average number of generated states and cells during planning, using \hbox{o-KPIECE} and p-KPIECE. The comparison between planning time of both planners using different levels of clutterness is shown in Fig.~\ref{fig:planningtime}. Since p-KPIECE considers uncertainty, it performs an additional step that is robustness evaluation of each generated state, which makes it computationally expensive as compared to the o-KPIECE.  As a result, for clutter scenes, the planning success rate of p-KPIECE is higher, as shown in Fig.~\ref{fig:successrate}. 
%\begin{figure}
%\begin{center}
%   \includegraphics[width=0.9\linewidth]{figures/particlesplanning.eps}
%   \caption{Graph of average planning time of 60 runs by varying the number of particle motions.}\label{fig:particlesplanning}
%\end{center}
%\end{figure}
The robustness evaluation depends on the number of particle motions that are used to compute the belief, the larger the number of particle motions the more robust the plan. However, 
the computational time increases with the number of particle motions, shown in Fig.~\ref{fig:particlesplanning}, and this may drop the planning success rate for not being able to find a solution on 
time, as shown in  Fig.~\ref{fig:planningsuccess}. Another parameter that may affect the planning process is the cell size used by KPIECE, that must be properly set. Big cell sizes do 
not ease the identification of the  relevant regions of the workspace and fail to guide the tree growth, resulting in long unrealistic paths. On the other hand, smaller cell sizes  better guide the 
search, thus enhancing the planning efficiency. Too small cell sizes, however, may increase the computational time required due to the huge number of cells generated. This effect on the planning time 
is depicted in Fig.~\ref{fig:cellsize}, where cell size is represented as a percentage of the side length with respect to the side of the cube that models the workspace. In this work 
the state space is projected to the workspace, using the end-effector position. 

%\begin{figure}
%\begin{center}
%   \includegraphics[width=0.9\linewidth]{figures/planningsuccessrate.eps}
%   \caption{Graph of average planning success rate of 60 runs by varying the number of particle motions.}\label{fig:planningsuccess}
%\end{center}
%\end{figure}

%\begin{figure}
%\begin{center}
%   \includegraphics[width=0.9\linewidth]{figures/cellsize2.eps}
%   \caption{Histogram of average planning time of 60 runs by varying the cell size as a percentage of the side length with respect to the side of the cube, representing the 
%workspace.}\label{fig:cellsize}
%\end{center}
%\end{figure}
\subsubsection{Comparison with Task Planning}
\begin{figure}
\begin{center}
   \includegraphics[width=1\linewidth]{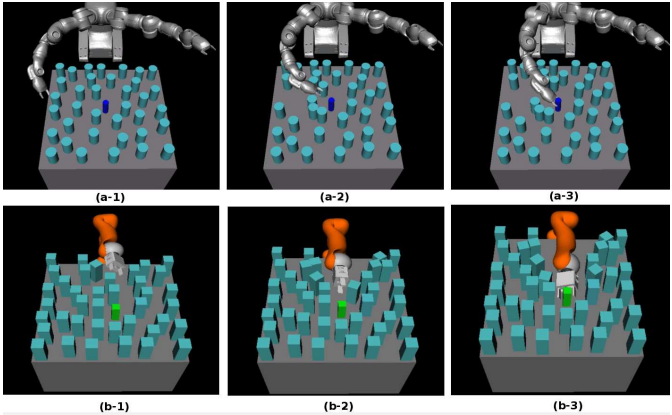}
   \caption{Sequence of the snapshot of the executions with the YuMi and the Kuka-LWR, for the comparison with the task and motion planning approach. Videos: 
https://goo.gl/ZorqCF and https://goo.gl/F4fZVL }\label{fig:TMP}
\end{center}
\end{figure}

The current proposal is also compared with the task and motion planning approaches (TMP) for grasping in the clutter without and with uncertainty ~\cite{hadfield2015,srivastava2014}, 
respectively. We have generated a qualitatively similar setup as that in those references, as shown in 
Fig.~\ref{fig:TMP}. Since the TMP approach presented in~\cite{srivastava2014} does not consider uncertainty in the environment, in order to be fair with the comparison, we have assumed that the 
uncertainty in the environment is almost negligible, and then one particle motion is enough to evaluate the robustness, i.e., $k=1$. Whereas, to compare with~\cite{hadfield2015}, the value of $k$ is 
set to 15. 
 The higher the value of $k$ the higher the confidence in 
the planner exploring safer regions, but the lower the computationally efficiency. The chosen value will depend on the problem; $k=15$  worked well in the example scenarios presented in this paper. 
Table~\ref{tablecomp} 
summarizes the results of the comparison. The data regarding the TMP approaches has been 
obtained from~\cite{srivastava2014} and ~\cite{hadfield2015}, being generated with an Intel Core i7-4770K machine with 16 GB RAM (i.e., a faster processor than the one used in the current study).  
The task and motion 
planning approach requires explicit reasoning to perform each action, i.e. repeatedly selects an object to pick, computes the collision free trajectory to grasp it, finds and appropriate placement 
location and moves the object to that location. In contrast, the current 
proposal does not require explicit reasoning about the complex dynamic multi-body interactions for moving in the clutter. At each step, the objects must satisfy the global set of constraints, that 
are easy to evaluate. This process makes it easy to compute the robust plan efficiently, even in the presence of uncertainty.   

\vspace{5mm}

\vspace{5mm}
\begin{table}
\centering
\begin{tabular}{|l|c|c|c|c|c|c|c|c|}  \hline
\multirow{3}{*}{Obj.}  & \multicolumn{4}{|c}{Success rate \%} & \multicolumn{4}{|c|}{Av. planning time (s)} \\  \cline{2-9}
  
   &TP    & TPU   &pKP &pKPU &TP    &TPU   &pKP    &pKPU \\ \hline
    5&-   &  95   &100 & 100 & -  & 89     & 2.89  & 9.88 \\ \hline
   10&-   &  95   &100 & 100 & -  & 162     &5.96  & 14.52\\ \hline
   15&100 &  83   &100 & 90  & 32 & 135     &9.15  & 29.43\\ \hline
   20&94  &  70   &100 & 86  & 57 & 229     &16.78 & 47.61\\ \hline
   25&90  &  68   &96  & 73  & 69 & 166     &26.09 & 72.28\\ \hline
   30&84  &  48   &90  & 66  & 77 & 122     &41.21 & 103.07\\ \hline
   35&67  &  -    &73  & 53  & 41 &  -    &49.62 & 134.94\\ \hline
   40&63  &  -    &60  & 40  & 68 &  -    &71.64 & 182.19\\ \hline
 \end{tabular}
 \caption{Comparison of p-KPIECE with task and motion planning approaches. Obj. represents the number of objects 
used, TP and TPU represent the approaches presented in~\cite{srivastava2014} and~\cite{hadfield2015}, respectively. pKP and pKPU represent the p-KPIECE with and without uncertainty, respectively.} 
\label{tablecomp}
 \end{table}

\subsection{Qualitative Analysis}
The quality of the solution path computed by p-KPIECE is improved as compared to the o-KPIECE. Since o-KPIECE does not evaluate the post-effect of dynamic interactions and the results of 
object-object interactions, that leads to the unreasonable results. For cluttered scenes, in most of the cases, it displaces the target object or drops the objects from the table surface, whereas 
p-KPIECE carefully evaluates the post-effect of dynamic interactions, interaction velocity and displacement covered by the object in the result of interactions, greatly increasing the planning success 
rate. Moreover, the incorporation of robustness evaluation phase results in the more stable and robust motion plans,
i.e. the integration of the results of robustness evaluation within the planning data structure allows the search to be guided  towards the regions that are safer,  
thus increasing the success in the final execution of the planned paths. 

The solution computed by p-KPIECE is also qualitatively different from the TMP solution. The TMP solution  is generated by integrating several move, pick and place actions. Each motion planning query 
is set to move the robot to a particular place (for picking or placing an object); these local queries do not consider the final goal. In contrast, p-KPIECE computes the solution in a more natural 
way, 
it launches a single query and moves away the objects obstructing the path.

Currently, no smoothing operation over the trajectory is applied, although it can easily be added by using deterministic control sampling strategies or by applying post processing over the computed 
path as presented in~\cite{van2011}. 

\section{CONCLUSIONS}\label{sec:conclusion}
This study proposes a randomized physics-based motion planner for planning the grasping motions in cluttered and unstructured environments. The developed framework takes into account the uncertainty 
in the objects' pose and in the contact dynamics. The KPIECE planner is enhanced by: a) introducing a motion sampler, for the extension of the tree, that samples motions and evaluates the belief 
about their 
robustness in the presence of uncertainty; b) biasing the tree exploration strategy based on the computed beliefs. The proposed planner enables single actions to move multiple objects, thereby 
avoiding the combinatorial explosion in the task planning part of TMP. The work is validated in simulation and in real environment against other approaches, such as ontological physic-based 
motion planner and task and motion planning approach. The results show significant advantages in terms of planning time, success rate and the quality of the computed solution path.    

\balance
\bibliographystyle{IEEEtran}%{ieeetr}{latex8}{plain}{unsrt}
\bibliography{ReferencesJR}

\end{document}